\documentclass[letterpaper]{article} 
\usepackage{aaai24}  
\usepackage{times}  
\usepackage{helvet}  
\usepackage{courier}  
\usepackage[hyphens]{url}  
\usepackage{graphicx} 
\usepackage{natbib}  
\usepackage{caption} 
\usepackage{algorithm}
\usepackage{algorithmic}

\usepackage{amsmath}
\usepackage{amssymb}
\usepackage{booktabs}
\usepackage{color}
\usepackage{multirow}
\usepackage{pifont}
\usepackage{url}
\usepackage{newfloat}
\usepackage{listings}
\DeclareCaptionStyle{ruled}{labelfont=normalfont,labelsep=colon,strut=off} 
\floatstyle{ruled}
\newfloat{listing}{tb}{lst}{}
\floatname{listing}{Listing}

\newcommand*\ie{\textit{i.e.}}
\title{M3SOT: Multi-frame, Multi-field, Multi-space 3D Single Object Tracking}
\author{
    Jiaming Liu\textsuperscript{\rm 1}, Yue Wu\textsuperscript{\rm 1}\thanks{Corresponding author.}, Maoguo Gong\textsuperscript{\rm 1}, Qiguang Miao\textsuperscript{\rm 1}, Wenping Ma\textsuperscript{\rm 1}, Can Qin\textsuperscript{\rm 2}
}
\affiliations{
    \textsuperscript{\rm 1}Xidian University, China \qquad
    \textsuperscript{\rm 2}Northeastern University, USA \\
    ljm@stu.xidian.edu.cn, \{ywu, qgmiao\}@xidian.edu.cn, gong@ieee.org, wpma@mail.xidian.edu.cn, qin.ca@northeastern.edu
}
\usepackage{bibentry}

\begin{document}

\maketitle

\begin{abstract}
	3D Single Object Tracking (SOT) stands a forefront task of computer vision, proving essential for applications like autonomous driving. Sparse and occluded data in scene point clouds introduce variations in the appearance of tracked objects, adding complexity to the task. In this research, we unveil M3SOT, a novel 3D SOT framework, which synergizes \textit{multiple} input frames (template sets), \textit{multiple} receptive fields (continuous contexts), and \textit{multiple} solution spaces (distinct tasks) in ONE model. Remarkably, M3SOT pioneers in modeling temporality, contexts, and tasks directly from point clouds, revisiting a perspective on the key factors influencing SOT. To this end, we design a transformer-based network centered on point cloud targets in the search area, aggregating diverse contextual representations and propagating target cues by employing historical frames. As M3SOT spans varied processing perspectives, we've streamlined the network—trimming its depth and optimizing its structure—to ensure a lightweight and efficient deployment for SOT applications. We posit that, backed by practical construction, M3SOT sidesteps the need for complex frameworks and auxiliary components to deliver sterling results. Extensive experiments on benchmarks such as KITTI, nuScenes, and Waymo Open Dataset demonstrate that M3SOT achieves state-of-the-art performance at 38 FPS. Our code and models are available at \url{https://github.com/ywu0912/TeamCode.git}.
	
\end{abstract}

\section{Introduction}
Visual object tracking is a basic task in computer vision, while single object tracking (SOT) is tracking a specific object in sequential data, considering only its initial pose. With the development of 3D sensors such as LiDAR, the acquisition of 3D data and the progress of 3D tasks become more active. In particular, great progress has been made in the 3D field based on point clouds \cite{wu2022centralized,wu2023self,wu2023correspondence,huang2023cross,liu2023exploring}. Yet, SOT remains challenging due to the variation in object appearance and the sparseness caused by sensors with inherent limitations.

Existing 3D SOT methods can be summarized into two main paradigms, \ie, Siamese network and spatio-temporal modeling. As a pioneering work, SC3D \cite{giancola2019leveraging} crops the target from the $(t-1)$-th frame and compares the target template with a large number of potential candidates in the $t$-th frame. P2B \cite{qi2020p2b} optimizes this process by taking the cropped template and the search area as inputs, and propagating the cues to the search area by training again to predict the current bounding box. This idea has broad implications for subsequent research. Yet, the paradigm rooted in Siamese networks overlooks the background information from two sequential frames. Moreover, it fails when the target is potentially absent in the current frame. To address these issues, M2-Track \cite{zheng2022beyond} presents a motion-centric approach, processing two point cloud frames as input and directly segmenting the target points from their respective backgrounds, eliminating the need for cropping. TAT \cite{lan2022temporal} ensures dependable target-specific feature propagation. It achieves this by sampling high-quality target templates derived from historical frames, applying template data across various timelines. However, these strategies predominantly operate on cropped subregions, which are fragmentary of essential contextual information in localization. Echoing the sentiments of CXTrack \cite{xu2023cxtrack}, leveraging the contextual information surrounding the target for predicting its current bounding box is indeed an applicable move.
\begin{figure}
	\centering
	\includegraphics[width=\linewidth]{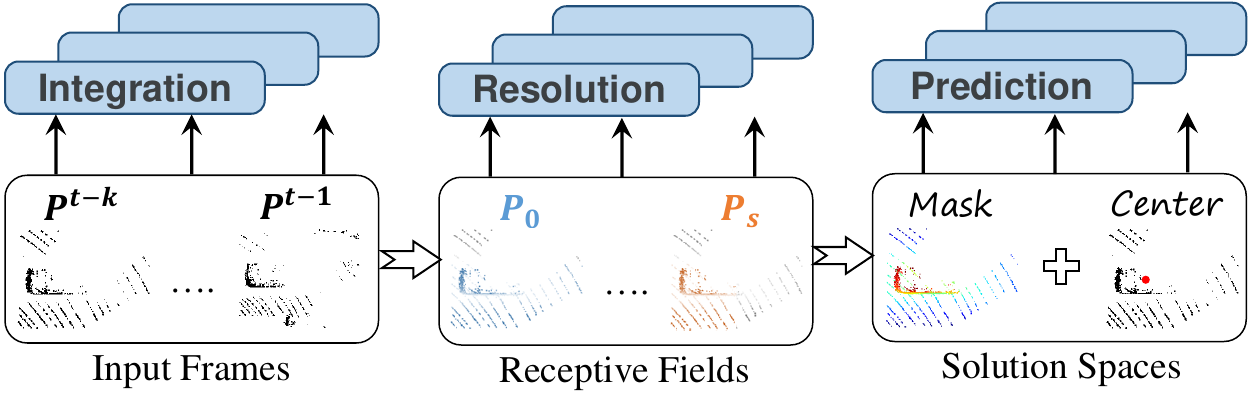}
	\vspace{-15pt}
	\caption{Illustration of the proposed M3SOT. M3SOT collects multi-frame point clouds for propagating target cues, and extracts spatio-temporal context information through multiple receptive fields. We set additional mask and center prediction tasks for the backbone at intermediate stages.}
	\vspace{-10pt}
	\label{fig1}
\end{figure}

Hence, a logical proposition might be: by \textit{integrating multi-frame input, motion modeling, and context extraction, would SOT performance enhance?} We refute this seemingly straightforward yet inelegant hypothesis. Through experimentation, we find that this approach places excessive strain on the network, potentially diminishing performance. Additionally, we re-examine the variables influencing SOT, unearthing three pivotal insights that bolster tracking.

1) \textit{Multiple input frames.} Directly using the template set composed of multi-frame point clouds \cite{lan2022temporal} and indirectly introducing motion modeling \cite{zheng2022beyond} or updating the memory library \cite{xu2023mbptrack} is of great significance for tracking, as the unique temporal nature of SOT tasks can play a significant role. Inspired by these, our key idea is simple, \ie, integrating past frames, gradually correcting errors and refining bounding boxes over time. Specifically, we employ a powerful attention mechanism to learn contexts from historical templates and then integrate them into the search area for rich information aggregation and precise object localization.

2) \textit{Multiple receptive fields.} Fusion of multi-scale features is a well-known technique. For 3D SOT, most methods tend to use PointNet++ \cite{qi2017pointnet++} or DGCNN \cite{wang2019dynamic} as the backbone for collecting multi-stage features. Yet, fusing these features is challenging, given the inherent tension between higher resolutions and expansive receptive fields. In response, we introduce a new multi-receptive field module with a transformer backbone designed to gather contextual information from multi-frame point clouds. Specifically, we obtain point cloud features representing the complete template through multi-stage computation-free range sampling and pointwise transformation. Our core insight is the conviction that predicting objects directly from sparse point features—without the risky truncation of the template—is both viable and effective \cite{chen2023voxelnext}.

3) \textit{Multiple solution spaces.} Reviewing the previous SOT journey, we find that most methods rely only on the final localization head to discriminate the bounding box after pointwise transformation \cite{hui20213d,zheng2021box,liu2023instance}. This paradigm is agnostic to the intermediate stages of the network under training, since only the point features with maximum probability are finally acquired. For this reason, we revisit SOT, whose discriminative process should be asymptotic, \ie, it can characterize the rough distribution of bounding boxes during the training process. To take full advantage of this cue, we set additional solution spaces in the intermediate stage for solving the mask and center of the predicted search area, with the former estimating the overall distribution of the bounding box and the latter pinpointing. 
Specifically, the designed transformer used to extract and transform point features has $L$ stacked layers, where the output of each layer is supervised, while only the updated search area features of the last layer are forwarded to the localization head for the prediction.


As a significant result, we achieve the framework unification and unleash the potential of 3D SOT. In short, we inherit the above three findings into a framework, M3SOT, as shown in Figure \ref{fig1}. M3SOT is reinvigorated in the loop with spatio-temporal cues in the input phase and contextual information and task reasoning in the intermediate phase. Benefiting from the information aggregation of historical templates, sufficient contextual information and additional hidden spaces, M3SOT can efficiently track specific targets even in the case of occlusion or missing. Extensive experiments show that M3SOT achieves state-of-the-art performance on three benchmarks while running at 38 FPS on a single NVIDIA RTX 3090 GPU.

\section{Related Work}
\noindent\textbf{3D SOT.} Recently, 3D point cloud-based tracking can effectively avoid problems such as reliance on RGB-D information and sensitivity to illumination changes and object size variations in the 2D image tracking domain. SC3D \cite{giancola2019leveraging} is the first 3D Siamese tracker based on shape completion that generates a large number of candidates in the search area and compares them with the cropped template, taking the most similar candidate as the tracking result. The pipeline relies on heuristic sampling and does not learn end-to-end, which is very time consuming. P2B \cite{qi2020p2b} addresses the previous problem by first using feature augmentation to enhance the perception of the specific template in the search area, and then using VoteNet \cite{qi2019deep} to localize the specific object in the search area. Most of the subsequent work basically follows the Siamese model. MLVSNet \cite{wang2021mlvsnet} aggregates information in multiple stages to achieve more effective target localization. BAT \cite{zheng2021box} introduces a box-aware module to enhance discriminative learning between object templates and search areas. V2B \cite{hui20213d} proposes a voxel-to-BEV object localization network, which projects sparse point features into a dense BEV feature map to address the sparsity of point clouds.

\noindent\textbf{3D SOT by Transformer.} Transformer \cite{vaswani2017attention} captures long-term dependencies of input sequences by the attention mechanism. Recently, transformer is applied to 3D vision and achieves good performance \cite{wu2023sacf,wu2023rornet,wu2023mpct,yuan2023egst,liu2023inter}. LTTR \cite{cui20213d}, PTTR \cite{zhou2022pttr}, and STNet \cite{hui20223d} introduce various attention mechanisms to 3D SOT tasks for better target-specific feature propagation. CXTrack \cite{xu2023cxtrack} uses adjacent frames and employs a target-centric transformer to propagate target cues into the current frame while exploring the contextual information around the target. This “tracking by attention” paradigm is on the rise, as it has been shown to be effective for interactive learning of templates and search areas. However, these methods only exploit the target cues in the latest frame while ignoring the rich information in the historical frames. Our proposed method is applied in this paradigm, but extends the temporal scope of existing methods. In particular, we demonstrate that joint past inference can provide robust representations of spatio-temporal objects to improve the tracking.

\noindent\textbf{3D SOT by Temporality.} Continuous temporal context with logical processes is meaningful for 3D cognition, especially for dynamic 3D SOT task. M2-Track \cite{zheng2022beyond} models consecutive frames as a motion-centric paradigm. TAT \cite{lan2022temporal} samples high-quality templates from historical frames and aggregate target cues. CAT \cite{gao2023spatio} aggregates the features of historical frames to enhance the representations of the templates. MBPTrack \cite{xu2023mbptrack} designs an external memory for historical frames, and propagates the tracked target clues from the memory to the current frame. Unlike them, we utilize the contextual information of historical frames to learn interactively with the current frame respectively. As our insight is simple and efficient: there is a comprehensive transformer and a task solver, just make sure the inputs are sufficient.

\section{Pilot Study: Revisit Multi-Frame 3D SOT}
\noindent\textbf{Problem Formulation.} In the 3D SOT task, given the initial bounding box (BBox) of the target in the first frame, the tracker aims at predicting the BBox of the target in the subsequent search area point cloud $\boldsymbol{P^t} \in \mathbb{R}^{N_t \times 3}$. It is generally assumed that the target size is fixed, and the rotation direction is just around the z-axis. Therefore, for each frame $\boldsymbol{P^t}$, the tracker only regresses the translational offsets $(\Delta x, \Delta y, \Delta z)$ and rotational angles $\Delta \theta$ from $\boldsymbol{P^{t-1}}$ to $\boldsymbol{P^t}$.

Further, the multi-frame 3D SOT extends the previous formulation, \ie, $\boldsymbol{P^{t-1}}$ becomes $\boldsymbol{P^{t-K:t-1}}$. In addition, to represent the position and pose of the tracked target on the historical frames, we utilize the predicted targeting masks as auxiliary inputs. As a result, we reformulate the 3D SOT as
\begin{equation}\label{eq1}
		\texttt{Track}(\boldsymbol{\{P, M\}^{t-K:t-1}}, \boldsymbol{P^t}) \mapsto (\Delta x, \Delta y, \Delta z, \Delta \theta).	
\end{equation}

\begin{figure}[]
	\centering
	\includegraphics[width=\linewidth]{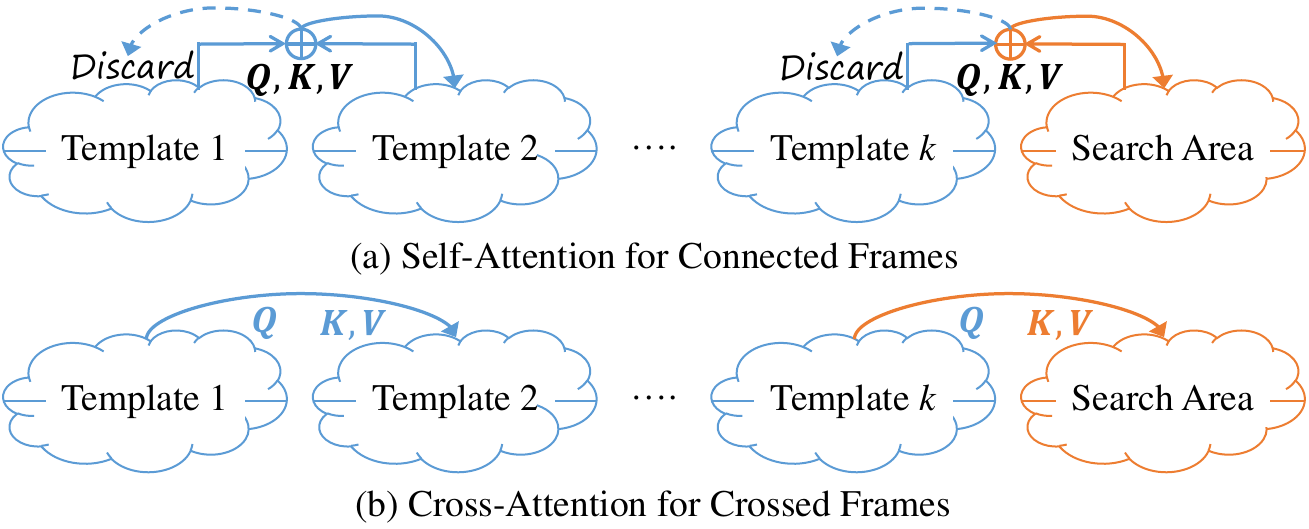}
	\vspace{-15pt}	
	\caption{Illustration of frame-by-frame target propagation. We verify two generative paradigms (a) and (b) by attention.} 
	\label{fig2}
\end{figure}

Since 3D SOT tracks the target in a dynamic sequence, it has timing. Therefore, we first discuss \textit{whether timing can be reflected by frame-by-frame propagation?} In other words, the template set is passed progressively from the first frame to the next frame to the final search area frame. We design two generative paradigms to study it, as shown Figure \ref{fig2}.

\noindent\textbf{(a) Self-attention.} We concatenate consecutive frames into a new frame and perform self-attention to split the next frame taking the cue propagation from the previous frame.

\noindent\textbf{(b) Cross-attention.} We transform the previous frame into a query matrix and perform cross-attention with the next frame to propagate cues to the next frame.

\begin{table}[]
	\centering
	\caption{Generative paradigms for frame-wise propagation.}
	\label{tab1}
	\vspace{-10pt}
	\resizebox{\linewidth}{!}{
		\begin{tabular}{l|cccc}
			\toprule
			Template set size (Frame) & 1 & 2 & 3 & 4\\
			\midrule
			(a) self-attention (Precesion) & 82.1 & 74.1\scriptsize($\downarrow$8.0) & 71.9\scriptsize($\downarrow$2.2) & 71.6\scriptsize($\downarrow$0.3) \\
			(b) cross-attention (Precesion) & 81.6 & 76.3\scriptsize($\downarrow$5.3) & 74.5\scriptsize($\downarrow$1.8) & 72.1\scriptsize($\downarrow$2.4) \\
			\bottomrule
	\end{tabular}}
	\vspace{-10pt}	
\end{table}

These two generative paradigms are negative for multi-frame 3D SOT (see Table \ref{tab1}, tested in KITTI Car). We conclude that the target clues in the template set cannot be propagated to the search area frame by frame, because the template sets originally have their own targets, and redundant propagation may make the search area get wrong signals.

\begin{figure}[]
	\centering
	\includegraphics[width=\linewidth]{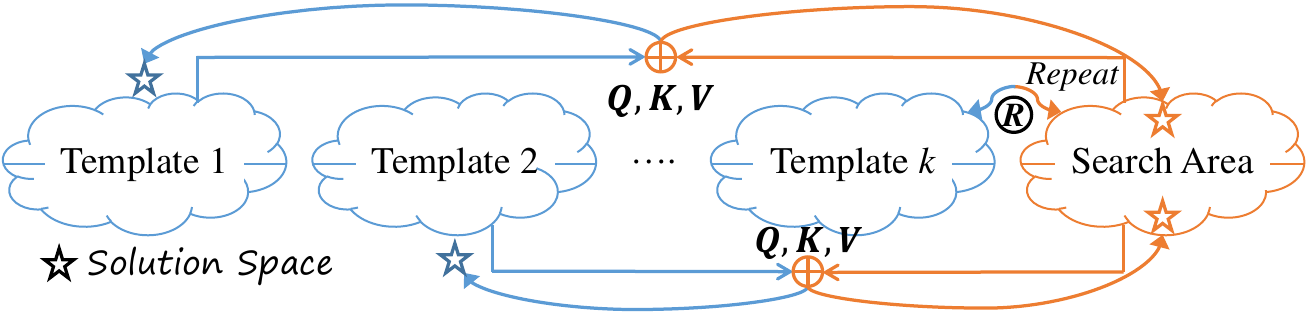}	
	\caption{Illustration of M3SOT's many-to-one target propagation. Novel solution spaces for intermediate predictions.} 
	\label{fig3}
	\vspace{-8pt}	
\end{figure}

Differently, our intuition is that discontinuous frames can be complementary. This is contrary to the above, as it is unnecessary to build potential movement in an unbalanced point cloud sequence. Recalling at the difficulties of 3D SOT, we argue that sparseness and occlusion are the most important factors. Therefore, we directly adopt the many-to-one matching scheme for 3D SOT, as shown in Figure \ref{fig3}.
\section{Proposed Method: M3SOT}
\begin{figure*}[t]
	\centering
	\includegraphics[width=0.95\linewidth]{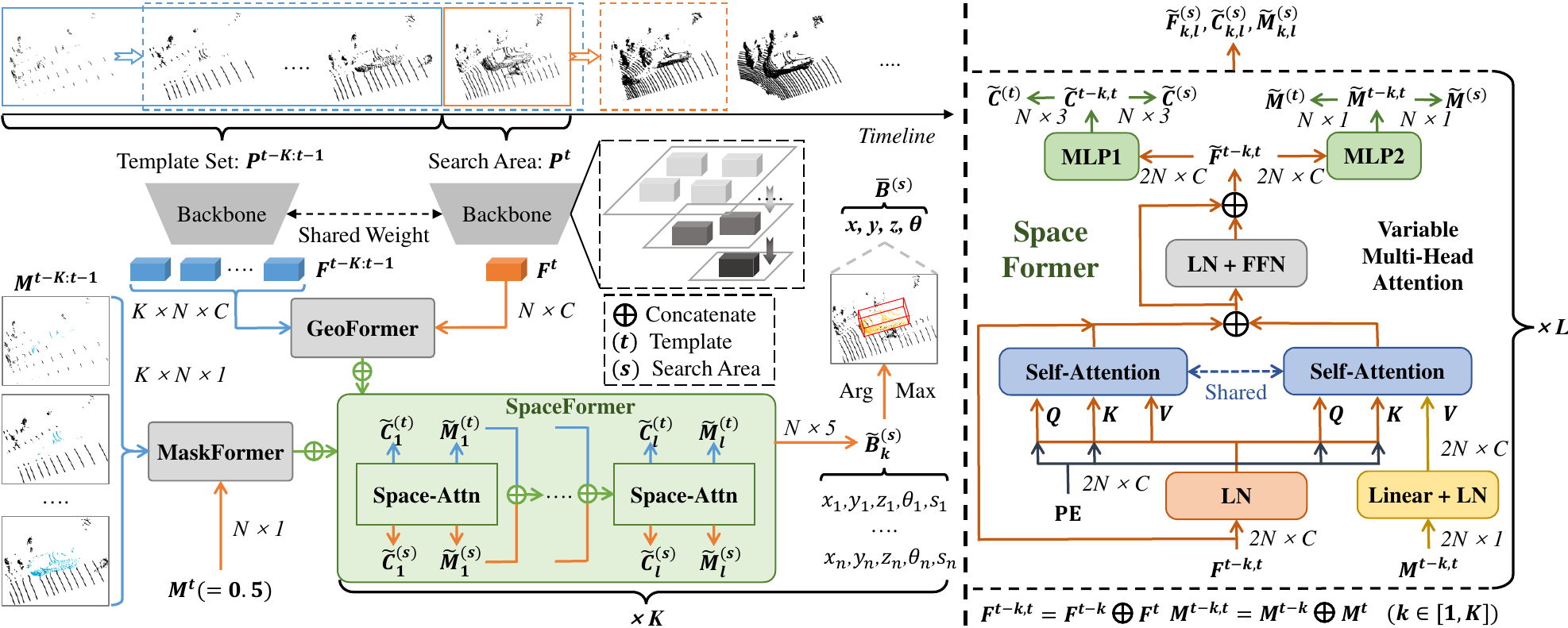}
	\caption{The overall framework of M3SOT. Given a point cloud sequence, M3SOT first employs a backbone with multiple receptive fields to extract the template set features of the previous $k$ frames and the search area features of the current frame. Then, the geometric and mask features obtained by GeoFormer and MaskFormer are fed to a multi-task SpaceFormer to explore the spatio-temporal context of consecutive frames and propagate the target cues of each template into the search area. In addition, we design a multi-layer network with variable multi-head attention in SpaceFormer to predict masks and centers.} 
	\label{fig4}
	\vspace{-8pt}
\end{figure*}
\noindent\textbf{Overview.} Based on Eq. (\ref{eq1}), we propose M3SOT, a multi-frame, multi-field, multi-space SOT framework to fully utilize the spatial and temporal information of history frames. The overall framework is shown in Figure \ref{fig4}, where the template set and search area are first divided based on a 3D input sequence $\{\boldsymbol{P^i}\}_{i=1}^t$. A shared backbone with a hierarchical structure is used to extract local geometric features of the point cloud and aggregate them into point features. $\boldsymbol{F^i} \in \mathbb{R}^{N \times C}$ denotes the point cloud features in the $i$-th frame, and the corresponding targetness mask $\boldsymbol{M^i} \in \mathbb{R}^{N \times 1}$ is obtained from the first frame or estimated from past frames to identify the tracked targets in historical frames. To unify the computation, we design a targetness mask $\boldsymbol{M^t}$ initialized to 0.5 for the current frame due to the consistent initial state of the unknown points. Inspired by \cite{xu2023cxtrack}, we concatenate the point features and targetness masks of the template set and the search area respectively to form a many-to-one pattern. Then, we design an interactive feature propagation module based on transformer to embed the geometric and mask information of the point cloud through GeoFormer and MaskFormer, respectively, and then feed them into SpaceFormer to perform spatial-separation tasks and feature transformations. Finally, the targeting proposals and confidences are predicted by the splited search area features with the support of localization head, and the proposal with the maximum confidence is identified as the bounding box.
\vspace{-5pt}	
\subsection{Start from Inputs: Multi-frame SOT}
Traditional Siamese trackers take as input a single template point cloud $\boldsymbol{P^{t-1}}$ and a search area point cloud $\boldsymbol{P^t}$, and match the closest target to the template within the search area. Differently, we focus on extracting rich temporal contextual information from a set of collected templates $\boldsymbol{P^{t-K:t-1}}$ for robust object localization. Considering the temporal context between the template set to the search area, we use simple and realistic closest frame sampling.

To exploit the potential of the template set to propagate cues to the search area, we construct multiple perceptual networks to predict the point-wise scores of the search area under the action of each historical template, and then select the point with the highest score for regression to the bounding box. Specifically, the historical templates and the search area are input to the network, where the former is used to provide reliable and necessary background information. Since different history templates are relevant to the search area in different degrees, the network can generate compatible regular terms for the weight of the search area with the support of shared weights. This paradigm empirically prevents the negative impact of low-quality templates on the search area.

\subsection{Refactor Point Features: Multi-field SOT}
Given that the 3D SOT task is realized by a point-specific transformation of the search area, it is important how to inject more prior information into the points and enhance their discriminative ability. One fact is that the target-specific features provided by the template set are the most critical clues. In addition, it is also beneficial to add the perceptual information of the BBox to the point. BAT \cite{zheng2021box} uses the 8 corners and 1 center of the point to the BBox as the additional information of the point. The difference is that we directly generate an additional mask set to record the probability of the point being in the BBox. These two factors can complement each other and are supervised differently.

We observe that using all points and masks directly as the only input results in two bad situations: 1) overloading the network and 2) poor and unstable results. This is due to the fact that not all points are equal, and the target points represent only a small fraction of the input points. Therefore, we design multiple receptive fields for the input point cloud, gradually decreasing the number and aggregating local information for points. Specifically, for the input point cloud $\boldsymbol{P_0}$, we generate new inputs $\boldsymbol{P_s}$ and $\boldsymbol{F_s}$ by a backbone with $\boldsymbol{S}$ range sampling and feature aggregation operations.
\begin{equation}\label{eq2}
	\boldsymbol{P_s} = RS(\boldsymbol{P_0}), \boldsymbol{F_s} = DGCNN(\boldsymbol{F_0}),
\end{equation}
where $RS$ requires no computation and retains the relationship between points, $DGCNN$ is used to extract point features with local aggregation \cite{wang2019dynamic}. 

Intuitively, deeper features are coarse but reliable since they gather more information through a larger receptive field. We generate the corresponding mask $\boldsymbol{M_s}$ through the sampled position indexes. Note that while range sampling may miss target points, background points aggregated with target points can yield robust predictions, which is an important inspiration for dealing with sparsity in point clouds.

\subsection{Integrate a Hybrid Transformer: Multi-space SOT}
To efficiently handle the template set and the search area, we aim to enhance both point features and localize a intra-frame target in the search area, while propagating target cues from historical frames to the current frame. Inspired by \cite{xu2023cxtrack}, we propose a hybrid transformer that integrates multiple inputs and tasks with consideration of timing.

\noindent\textbf{MaskFormer.} To fully utilize the predicted results of history frames, we encode the point-box relationships of the template point clouds in a masked manner, \ie, $\mathbf{ME}$. Note that the mask of the $i$-th point $p_i$ is defined as
\begin{equation}\label{eq3}
	m^{(t)}_i=\{
        \begin{array}{ll}
		0, & p^{(t)}_i \text { not in } \boldsymbol{B}^{(t)}, \\
		1, & p^{(t)}_i \text { in } \boldsymbol{B}^{(t)}.
	\end{array}
\end{equation}

$\mathbf{ME}$ is similar to the positional encoding $\mathbf{PE}$, and $N$ here denotes the number of sampled points. In addition, we set a mask initialized to 0.5 on the search area for computation.

\noindent\textbf{GeoFormer.} As the template set and the search area are processed by the backbone in a many-to-one manner, the extracted geometric features $\boldsymbol{F^{t-k,t}} = \boldsymbol{F^{t-k}} \oplus \boldsymbol{F^{t}}$ are sufficient to represent the overall information of the two, where $k$ represents the $k$-th template in front of the search area. 

\noindent\textbf{SpaceFormer.} To explore how point features predict bounding boxes, we feed $\boldsymbol{F^{t-k,t}}$ and $\boldsymbol{M^{t-k,t}}$ to the space-attention module in SpaceFormer, since both inputs are included, the cross-attention is potentially performed. This process is the cornerstone of delivering the target cues of the template set to the search area, as shown in Figure \ref{fig4} (right). 

Specifically, we first employ $LN(\cdot)$ \cite{ba2016layer} to normalize features, which is formulated as
\begin{equation}
	\boldsymbol{\widetilde{F}^{t-k,t}} = LN(\boldsymbol{F^{t-k,t}}).
\end{equation}

Then, we build the basic components of attention: query $\boldsymbol{F_Q} \in \mathbb{R}^{2N \times C}$, key $\boldsymbol{F_K} \in \mathbb{R}^{2N \times C}$ and value $\boldsymbol{F_V} \in \mathbb{R}^{2N \times C}$, and add the positional encoding ($\mathbf{PE}$) to the query and key.
\begin{equation}
	\boldsymbol{Q} = \boldsymbol{K} = \boldsymbol{\widetilde{F}^{t-k,t}} + \mathbf{PE}, \boldsymbol{V} = \boldsymbol{\widetilde{F}^{t-k,t}}.
\end{equation}

Importantly, SpaceFormer employs a global multi-head self-attention module to model dependencies between point and mask features, formulated as
\begin{equation}
	\begin{split}
		\boldsymbol{\widetilde{F}^{t-k,t}} = \boldsymbol{F^{t-k,t}} &+{MHA}(\boldsymbol{Q},\boldsymbol{K},\boldsymbol{V}) \\ &+ {MHA}(\boldsymbol{Q},\boldsymbol{K},\mathbf{ME}),
	\end{split}
\end{equation}
where $MHA$ stands for multi-head attention, and the single-head attention with $d_{h} = C / H$ of the $i$-th in all subspaces being concatenated is $\boldsymbol{Q_i}$, $\boldsymbol{K_i}$, $\boldsymbol{V_i}$, calculated as
\begin{equation}
	{Attn}(\boldsymbol{Q_i}, \boldsymbol{K_i}, \boldsymbol{V_i})={softmax}\left(\frac{\boldsymbol{Q_i} \boldsymbol{K_i}^T}{\sqrt{d_{h}}}\right) \boldsymbol{V_i}.
\end{equation}

One reference is CXTrack which sets the number of layers to $L = 4$, the number of heads to $H = 1$. However, we argue that the gain this brings to multi-frame SOT is limited, since using the same configuration for different templates makes it difficult to model the network dynamically. Therefore, we propose a variable multi-attention mechanism that is simple and effective. Briefly, for different templates $\boldsymbol{P_k^{(t)}}$, we set the depth $L$ of the network to be proportional to $H$ for obtaining $\boldsymbol{\widetilde{F}_{k,l}^{(t)}}$, and the same for $\boldsymbol{\widetilde{F}_{k,l}^{(s)}}$.

For $\boldsymbol{\widetilde{F}_{k,l}^{t-k,t}}$ generated by different inputs at different layers, we separate them into $\boldsymbol{\widetilde{F}_{k,l}^{(t)}}$ and $\boldsymbol{\widetilde{F}_{k,l}^{(s)}}$. Our insight is that setting supervision on the outputs of each layer enables the targeting masks and centers to be consistently refined, 
\begin{gather}
	\boldsymbol{\widetilde{F}^{t-k,t}_{l}} = \boldsymbol{\widetilde{F}^{t-k,t}_{l-1}} + FFN(LN(\boldsymbol{\widetilde{F}^{t-k,t}_{l-1}})), \\
	\resizebox{.9\hsize}{!}{$
		\boldsymbol{\widetilde{M}_{k,l}^{(t,s)}} = MLP_{m}(\boldsymbol{\widetilde{F}^{t-k,t}_{l}}), \boldsymbol{\widetilde{C}_{k,l}^{(t,s)}} = MLP_{c}(\boldsymbol{\widetilde{F}^{t-k,t}_{l}}),$}
\end{gather}
where ${FFN}(x)=\operatorname{ReLU}\left(x W_1+b_1\right) W_2+b_2$.

Finally, $\boldsymbol{\widetilde{F}_{k,l}^{(s)}}$, $\boldsymbol{\widetilde{C}_{k,l}^{(s)}}$, and $\boldsymbol{\widetilde{M}_{k,l}^{(s)}}$ in the last layer are forwarded to X-RPN \cite{xu2023cxtrack} to predict the BBox,
\begin{equation}
	\boldsymbol{\widetilde{B}_{k}^{(s)}} = X\!-\!RPN(\boldsymbol{\widetilde{F}_{k,l}^{(s)}}, \boldsymbol{\widetilde{C}_{k,l}^{(s)}}, \boldsymbol{\widetilde{M}_{k,l}^{(s)}}).
\end{equation}

Since there are $\boldsymbol{K}$ templates, there are $\boldsymbol{K}$ versions of the search area, and we concatenate them to predict the BBox $\boldsymbol{\bar{B}^{(s)}}$ with the maximum confidence score.

\section{Experiments}
\subsection{Experimental Settings}
\noindent\textbf{Datasets.} We compare the proposed M3SOT with state-of-the-art methods on three large datasets: KITTI \cite{geiger2012we}, nuScenes \cite{caesar2020nuscenes}, and Waymo OpenDataset (WOD) \cite{sun2020scalability}. Following \cite{hui20213d,pang2021model}: For KITTI, we divide the training sequence into three parts, 0-16 for training, 17-18 for validation, and 19-20 for testing. For the more challenging nuScenes, we use its validation split to evaluate our model, which contains 150 scenarios. For WOD, we evaluate our method on 1121 tracklets, which is categorized into easy, medium, and difficult parts based on the sparsity.	

\noindent\textbf{Implementation Details.} We dilate the ground truth BBox by 2 meters to track possible objects in the area. DGCNN \cite{wang2019dynamic} with different configurations is used as the feature extractor, and X-RPN \cite{xu2023cxtrack} with the same parameters is used as the localization head.

\noindent\textbf{Evaluation Metrics.} We follow One Pass Evaluation (OPE) \cite{kristan2016novel}. For both predicted and ground truth BBoxes, \textit{Success} measures the intersection over union (IOU) between the two BBoxes from 0 to 1, while \textit{Precision} measures the area under curve (AUC) for the distance between their centers from 0 to 2 meters.

\subsection{Experimental Results}
\begin{table}[t]
	\caption{Comparison with the SOTA methods on KITTI. “Mean” denotes the average results weighted by frame numbers. \textbf{Bold} and \underline{underline} represent the best and second best results, respectively. Success/Precision are reported.}
	\vspace{-5pt}
	\label{tab2}
	\centering
	\resizebox{\linewidth}{!}{
		\begin{tabular}{c|cccc|c}
			\toprule
			\multirow{2}*{Method} &  Car& Pedestrian& Van&  Cyclist& Mean\\
			& 6424& 6088& 1248& 308& 14068\\
			\midrule
			SC3D & 41.3/57.9 & 18.2/37.8 & 40.4/47.0 & 41.5/70.4& 31.2/48.5\\
			P2B & 56.2/72.8 & 28.7/49.6 & 40.8/48.4 & 32.1/44.7 & 42.4/60.0\\
			LTTR & 65.0/77.1 & 33.2/56.8 & 35.8/45.6 & 66.2/89.9 & 48.7/65.8\\
			MLVSNet & 56.0/74.0 & 34.1/61.1 & 52.0/61.4 & 34.3/44.5 & 45.7/66.7\\
			BAT & 60.5/77.7 & 42.1/70.1 & 52.4/67.0 & 33.7/45.4& 51.2/72.8\\
			PTT & 67.8/81.8& 44.9/72.0 & 43.6/52.5 & 37.2/47.3 & 55.1/74.2\\
			V2B & 70.5/81.3 & 48.3/73.5 & 50.1/58.0 & 40.8/49.7 & 58.4/75.2\\
			CMT & 70.5/81.9 & 49.1/75.5 & 54.1/64.1 & 55.1/82.4 & 59.4/77.6\\
			PTTR & 65.2/77.4 & 50.9/81.6& 52.5/61.8 & 65.1/90.5 & 58.4/77.8\\
			CAT & 66.6/81.8 & 51.6/77.7 & 53.1/69.8 & 67.0/90.1 & 58.9/79.1\\
			STNet & 72.1/84.0 & 49.9/77.2 & 58.0/70.6 & 73.5/93.7 & 61.3/80.1\\
			TAT & 72.2/83.3 & 57.4/84.4 & 58.9/69.2 & \underline{74.2}/93.9 & 64.7/82.8\\
			M2-Track & 65.5/80.8 & 61.5/88.2 & 53.8/70.7 & 73.2/93.5 & 62.9/83.4\\
			CXTrack & 69.1/81.6 & \underline{67.0}/91.5 & \underline{60.0}/71.8 & \underline{74.2}/\textbf{94.3} & 67.5/85.3\\
			MBPTrack & \underline{73.4}/\underline{84.8} & \textbf{68.6}/\textbf{93.9} & \textbf{61.3}/\underline{72.7} & \textbf{76.7}/\textbf{94.3} & \textbf{70.3}/\underline{87.9} \\
			\midrule		
			M3SOT & \textbf{75.9}/\textbf{87.4} & 66.6/\underline{92.5} & 59.4/\textbf{74.7} & 70.3/93.4 & \textbf{70.3}/\textbf{88.6} \\
			\textit{Improvement} & \textcolor{green}{$\uparrow$2.5}/\textcolor{green}{$\uparrow$2.6} & \textcolor{red}{$\downarrow$2.0}/\textcolor{red}{$\downarrow$1.4} & \textcolor{red}{$\downarrow$0.9}/\textcolor{green}{$\uparrow$2.0} & \textcolor{red}{$\downarrow$6.4}/\textcolor{red}{$\downarrow$0.9} &  0.0/\textcolor{green}{$\uparrow$0.7} \\
			\bottomrule	
	\end{tabular}}
	\vspace{-5pt}
\end{table}

\noindent\textbf{Evaluation on KITTI.} We perform a comprehensive comparison of our M3SOT with previous state-of-the-art methods on the KITTI dataset, including SC3D \cite{giancola2019leveraging}, P2B \cite{qi2020p2b}, LTTR \cite{cui20213d}, MLVS-Net \cite{wang2021mlvsnet}, BAT \cite{zheng2021box}, PTT \cite{shan2021ptt}, V2B \cite{hui20213d}, CMT \cite{guo2022cmt}, PTTR \cite{zhou2022pttr}, STNet \cite{hui20223d}, TAT \cite{lan2022temporal}, M2-Track \cite{zheng2022beyond}, CXTrack \cite{xu2023cxtrack} and MBPTrack \cite{xu2023mbptrack}. As shown in Table \ref{tab2}, M3SOT performs excellently overall and outperforms the recent CXTrack and MBPTrack. Note that, in order to standardize the training setup, the reported M3SOT is based on a template set of size 2. However, the dependence on the number of history frames varies across categories, see the subsequent ablation experiments. Compared to TAT and MBPTrack, which also utilize history frames, we tap the following advantages of M3SOT: 1) Unlike TAT, which considers complex sampling and aggregation operations for the template set, M3SOT only requires simple many-to-one matching; 2) Unlike MBPTrack, which focuses on changing only the BBox, M3SOT filters the BBox under the action of historical templates on the search area. As a result, the well-thought-out and elegant M3SOT is more suitable for 3D SOT.
\begin{figure*}[t]
	\centering
	\includegraphics[width=\linewidth]{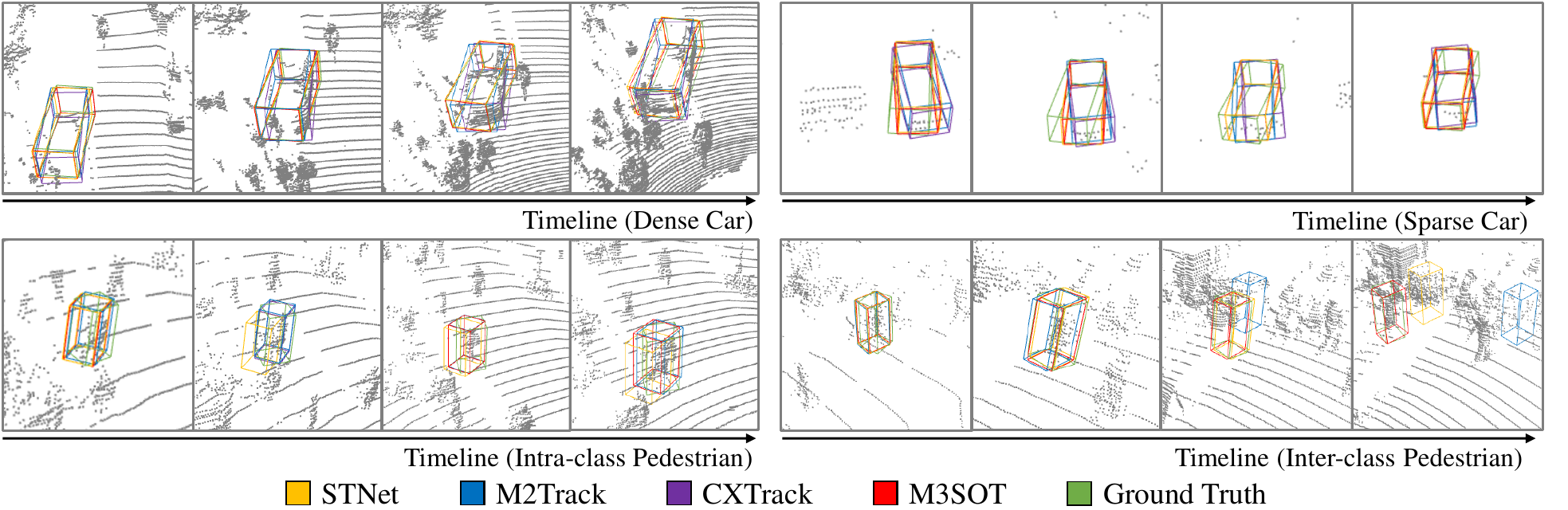}
	\vspace{-15pt}
	\caption{Visualization of tracking results from different methods on KITTI.} 
	\label{fig5}
	\vspace{-8pt}
\end{figure*}

We visualize the tracking results on KITTI, as shown in Figure \ref{fig5}. For Cars, the spatio-temporal context from historical frames allows M3SOT to produce discriminative semantic perceptions for the search area compared to non-multi-frame methods. For Pedestrians, most methods are prone to localize the wrong target due to the changing appearance of the target and distractors. However, due to the full use of temporal information, our M3SOT is able to accurately track the target in the presence of occlusions and appearance changes, and the aggregated information is more richer.

\noindent\textbf{Evaluation on nuScenes and WOD.} To validate the generalization ability of M3SOT, we follow \cite{hui20213d,pang2021model} and test the trained model on nuScenes and WOD. Note that the KITTI and WOD data are captured by 64-beam LiDAR, while the nuScenes data are captured by 32-beam LiDAR. Therefore, it is more challenging to generalize the trained model on the nuScenes dataset.
\begin{table}[]
	\caption{Comparison with the SOTA methods on nuScenes.}
	\label{tab3}
	\vspace{-8pt}
	\centering
	\resizebox{\linewidth}{!}{
		\begin{tabular}{c|cccc|c}
			\toprule
			\multirow{2}*{Method} &  Car& Pedestrian& Truck&  Bicycle& Mean\\
			& 15578& 8019 & 3710 & 501 & 27808\\
			\midrule
			SC3D & 25.0/27.1 & 14.2/16.2 & 25.7/21.9 & 17.0/18.2& 21.8/23.1\\
			P2B & 27.0/29.2 & 15.9/22.0 & 21.5/16.2 & 20.0/26.4 & 22.9/25.3\\
			BAT & 22.5/24.1 & 17.3/24.5 & 19.3/15.8 & 17.0/18.8& 20.5/23.0\\
			V2B & 31.3/35.1 & 17.3/23.4 & 21.7/16.7 & \textbf{22.2}/19.1 & 25.8/29.0\\
			STNet & 32.2/36.1 & 19.1/27.2 & 22.3/16.8 & \underline{21.2}/\textbf{29.2} & 26.9/30.8\\
			CXTrack & 29.6/33.4 & 20.4/32.9 & 27.6/20.8 & 18.5/26.8 & 26.5/31.5\\
			\midrule		
			M3SOT-F1 & \textbf{34.9}/\textbf{39.9} & \underline{23.3}/25.6 & \textbf{30.4}/\textbf{27.0} & 16.5/22.6 & \textbf{30.6}/33.7 \\
			\textbf{M3SOT-F2} & \underline{34.2}/\underline{38.6} & \textbf{24.6}/\textbf{37.8} & \underline{29.6}/\underline{25.5} & {18.8}/\underline{27.9} & \underline{30.5}/\textbf{36.4} \\
			M3SOT-F3 & 33.7/38.3 & 22.2/\underline{34.0} & 26.4/23.0 & 18.7/25.8 & 29.1/\underline{34.8} \\
			M3SOT-F4 & 32.4/36.8 & 21.7/32.8 & 28.0/23.8 & 17.0/21.6 & 28.4/33.6 \\
			\bottomrule	
	\end{tabular}}
	\vspace{-5pt}
\end{table}

\begin{figure}[t]
	\centering
	\includegraphics[width=\linewidth]{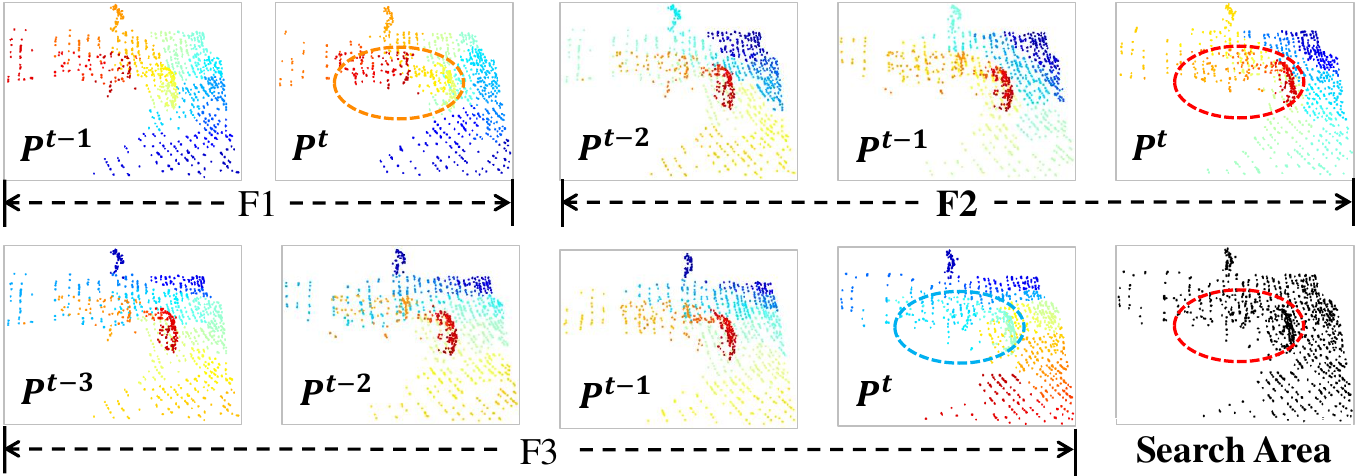}	
	\vspace{-15pt}	
	\caption{The effect of template set size on the search area.} 
	\label{fig6}
	\vspace{-8pt}	
\end{figure}

We set up four variants for M3SOT, each using models trained on the template set of sizes from 1 to 4. As shown in Table \ref{tab3}, our method achieves SOTA performance on the nuScenes, comprehensively outperforming previous methods. As a conclusion, M3SOT can not only generalize across different datasets, but also choose different configurations for different scenarios. Further, we visualize the impact of different template sets on the results in Figure \ref{fig6} to explore how the cross-domain model aggregates features and predicts semantics in a new domain. It is observed that different template sets have different impacts on the search area. Like KITTI with different densities, the template point clouds in the F2 are sufficient to propagate valid and complementary target cues to the search area point cloud, and too many or too few templates are detrimental to feature propagation.
\begin{table*}[]
	\caption{Comparison with the SOTA methods on Waymo Open Dataset.}
	\vspace{-8pt}
	\label{tab4}
	\centering
	\resizebox{\linewidth}{!}{
		\setlength\tabcolsep{1mm}{
			\begin{tabular}{c|cccc|cccc|c}
				\toprule
				\multirow{2}*{Method} &  \multicolumn{4}{|c|}{Vehicle} & \multicolumn{4}{|c|}{Pedestrian} & \multirow{2}*{Mean(427483)}\\
				& Easy(67832) & Medium(61252) & Hard(56647) & Mean(185731) & Easy(85280) & Medium(82253) & Hard(74219) & Mean(241752) &\\
				\midrule
				P2B & 57.1/65.4 & 52.0/60.7 & 47.9/58.5 & 52.6/61.7 & 18.1/30.8 & 17.8/30.0 & 17.7/29.3 & 17.9/30.1 & 33.0/43.8\\
				BAT & 61.0/68.3 & 53.3/60.9 & 48.9/57.8 & 54.7/62.7 & 19.3/32.6 & 17.8/29.8 & 17.2/28.3 & 18.2/30.3 & 34.1/44.4\\
				V2B & 64.5/71.5 & 55.1/63.2 & 52.0/62.0 & 57.6/65.9 & 27.9/43.9 & 22.5/36.2 & 20.1/33.1 & 23.7/37.9 & 38.4/50.1\\
				STNet & 65.9/72.7 & 57.5/66.0 & 54.6/64.7 & 59.7/68.0 & 29.2/45.3 & 24.7/38.2 & 22.2/35.8 & 25.5/39.9 & 40.4/52.1\\
				TAT & 66.0/72.6 & 56.6/64.2 & 52.9/62.5 & 58.9/66.7 & 32.1/49.5 & 25.6/40.3 & 21.8/35.9 & 26.7/42.2 & 40.7/52.8 \\
				CXTrack & 63.9/71.1 & 54.2/62.7 & 52.1/63.7 & 57.1/66.1 & 35.4/55.3 & 29.7/47.9 & 26.3/44.4 & 30.7/49.4 & 42.2/56.7\\
				M2Track & 68.1/75.3 & \underline{58.6}/66.6 & 55.4/64.9 & 61.1/69.3 & 35.5/54.2 & 30.7/48.4 & 29.3/45.9 & 32.0/49.7 & 44.6/58.2\\
				MBPTrack & \underline{68.5}/\underline{77.1} & 58.4/\underline{68.1} & \underline{57.6}/\underline{69.7} & \underline{61.9}/\underline{71.9} & \textbf{37.5}/\textbf{57.0} & \textbf{33.0}/\textbf{51.9} & 30.0/48.8 & \textbf{33.7}/\textbf{52.7} & \underline{46.0}/\underline{61.0}\\
				\midrule		
				M3SOT & \textbf{70.4}/\textbf{79.6} & \textbf{60.7}/\textbf{70.6} & \textbf{61.5}/\textbf{73.3} & \textbf{64.5}/\textbf{74.7} & \underline{36.3}/\underline{56.2} & \underline{31.6}/\underline{50.7} & \textbf{30.1}/\textbf{48.9} & \underline{32.8}/\underline{52.1} & \textbf{46.6}/\textbf{61.9} \\
				\textit{Improvement} & \textcolor{green}{$\uparrow$1.9}/\textcolor{green}{$\uparrow$2.5} & \textcolor{green}{$\uparrow$2.1}/\textcolor{green}{$\uparrow$2.5} & \textcolor{green}{$\uparrow$3.9}/\textcolor{green}{$\uparrow$3.6} & \textcolor{green}{$\uparrow$2.6}/\textcolor{green}{$\uparrow$2.8} &  \textcolor{red}{$\downarrow$1.2}/\textcolor{red}{$\downarrow$0.8} & \textcolor{red}{$\downarrow$1.2}/\textcolor{red}{$\downarrow$1.4} & \textcolor{green}{$\uparrow$0.1}/\textcolor{green}{$\uparrow$0.1} & \textcolor{red}{$\downarrow$0.9}/\textcolor{red}{$\downarrow$0.6} & \textcolor{green}{$\uparrow$0.6}/\textcolor{green}{$\uparrow$0.9} \\
				\bottomrule	
	\end{tabular}
            }
        }
	\vspace{-8pt}
\end{table*}

The comparison results on WOD are shown in Table \ref{tab4}. At different sparsity levels, our method is competitive than other methods, with an average gain of +0.6\%/+0.9\% compared to the recent MBPTrack. In any case, our M3SOT can not only accurately track a variety of targets, but also can be effectively generalized to unseen scenarios.

\subsection{Ablation Studies}
To verify the effectiveness of “M3” in M3SOT, we conduct ablation studies on the KITTI. In particular, “All” means that all categories are trained, not the weighted results above.
\begin{table}[]
	\caption{Ablation studies: template set sizes.}
	\vspace{-8pt}
	\label{tab5}
	\centering
	\resizebox{\linewidth}{!}{
		\begin{tabular}{c|cccc|c}
			\toprule
			Frame &  Car& Pedestrian& Van&  Cyclist& All\\
			\midrule
			{\textit{K}} = 1 &  72.8/84.7 & 66.2/91.1 & \textbf{63.1}/\textbf{78.0}&  69.6/92.5 & \textbf{70.1}/88.7\\
			\textbf{{\textit{K}} = 2} &  \textbf{75.9}/\textbf{87.4} & \textbf{66.6}/\textbf{92.5} & 59.4/74.7 & 70.3/93.4 & 70.0/\textbf{88.9}\\
			{\textit{K}} = 3 &  73.5/85.3 & 64.8/90.8 & 58.9/74.4&  \textbf{72.0}/\textbf{93.6} & 68.7/87.7\\
			{\textit{K}} = 4 &  71.8/83.1 & 62.9/88.3 & 62.5/76.2&  71.7/93.3 & 68.5/86.9\\
			\bottomrule	
	\end{tabular}}
	\vspace{-5pt}
\end{table}

\noindent\textbf{Multi-frame SOT.} To explore the effect of template set size on the propagation of target cues in the search area, we report the results in Table \ref{tab5}. When the template set size is set to 1, only the previous frame $\boldsymbol{P^{t-1}}$ is used to train and test our model. M3SOT in this case can be regarded as a Siamese-based network. We argue that the targets of the Van category can be tracked well in this state, which is related to its less deformation and few intraclass interferers. A basic phenomenon is that different template set sizes have certain effects on the tracking performance of different categories. Based on this, we observe that performance starts to degrade when tracking more than 2 frames, since too many historical frames allow the network to collect redundant features and backfire output spaces. Compared with MBPTrack using 4 frames and TAT using 8 frames, our M3SOT can achieve superior results at a lower cost (with default 2 frames). 
\begin{table}[]
	\caption{Ablation studies: feature generation ways.}
	\vspace{-8pt}
	\label{tab6}
	\centering
	\resizebox{\linewidth}{!}{
		\begin{tabular}{c|cccc|c}
			\toprule
			Ratio &  Car& Pedestrian& Van&  Cyclist& All\\
			\midrule
			{[1]} & 59.2/73.0 & 51.7/89.2 & 28.4/28.6 & 33.3/82.7 & 54.8/76.6\\
			{[2]} & 60.6/75.8 & 52.1/89.1 & 50.4/56.8 & 22.5/42.9 & 57.3/77.5\\
			{[2,4]} & 70.4/81.8 & 53.9/87.5 & 53.4/60.4 & 39.7/85.0 & 64.5/83.7\\
			\textbf{[2,4,8]} & \textbf{75.9}/\textbf{87.4} & \textbf{66.6}/\textbf{92.5} & \textbf{59.4}/\textbf{74.7} & \textbf{70.3}/\textbf{93.4} & \textbf{70.0}/\textbf{88.9}\\
			{[2,4,8,16]} & 71.6/83.5 & 58.2/88.9 & 58.3/66.2 & 22.3/38.1 & 64.4/83.5\\
			\bottomrule	
	\end{tabular}}
	\vspace{-5pt}
\end{table}

\noindent\textbf{Multi-field SOT.} DGCNN is used as the backbone to extract features, and we further discuss how to provide point features for the 3D SOT task. We set five receptive fields for the input point cloud according to different point sampling and feature aggregation: “Ratio: [1]” means that all points are selected, and the feature aggregation operation of k-nearest neighbors is performed; the other four settings are performed in equal proportions 1/2 sampling and pointwise aggregation operations. As shown in Table \ref{tab6}, we can observe that if all or half of the points after aggregated local features are used as input, the effect is not satisfactory. We argue that in the process of predicting unstable objects in sparse point clouds, the majority of points existing in shallow networks is noisy, and a large number of background points and target points cannot be distinguished. In addition, it is impossible to achieve good results only by extracting point features that may be less than the target point. Therefore, we choose a three-layer EdgeConv with 128 points, which can be used as the optimal carrier to represent targets.
\begin{table}[t]
	\caption{Ablation studies: intermediate space tasks.}
	\vspace{-8pt}
	\label{tab7}
	\centering
	\resizebox{\linewidth}{!}{
		\begin{tabular}{cc|cccc|c}
			\toprule
			M & C & Car& Pedestrian& Van&  Cyclist& All\\
			\midrule
			&  & 24.4/32.3 & 36.7/67.0 & 16.7/21.0 & 63.6/90.1 & 25.7/41.0\\
			\ding{51} &  & 72.6/83.8 & 65.3/90.5 & 57.5/69.5 & 70.2/93.1 & 69.9/88.0\\
			& \ding{51} &  74.5/85.7 & 61.5/87.7 & 57.6/69.4 & \textbf{72.3}/\textbf{93.5} & \textbf{70.1}/88.3\\
			\ding{52} & \ding{52} & \textbf{75.9}/\textbf{87.4} & \textbf{66.6}/\textbf{92.5} & \textbf{59.4}/\textbf{74.7} & 70.3/93.4 & 70.0/\textbf{88.9}\\
			\bottomrule	
	\end{tabular}}
	\vspace{-5pt}
\end{table}

\noindent\textbf{Multi-space SOT.} We also investigate the effectiveness of mask prediction and center regression in SpaceFormer, which are used as intermediate tasks and supervised. Following CXTrack, we replace the mask and center layers present in the prediction space with classical transformer layers to integrate the targeting information, and analyze the effect of direct regression to the bounding boxes. As shown in Table \ref{tab7}, if both are eliminated, the performance is quite bad, even heavier than CXTrack's results. We infer that the point features with multiple frames and multiple fields are extremely rich and need to be analyzed progressively through a hierarchical network, requiring a shallow-to-depth process. In addition, the absence of mask or center makes M3SOT only slightly degraded, reflecting the inclusiveness of our method to them, \ie, there is no need for excessive or complex tasks in the intermediate space.
\begin{figure}[]
	\centering
	\includegraphics[width=0.96\linewidth]{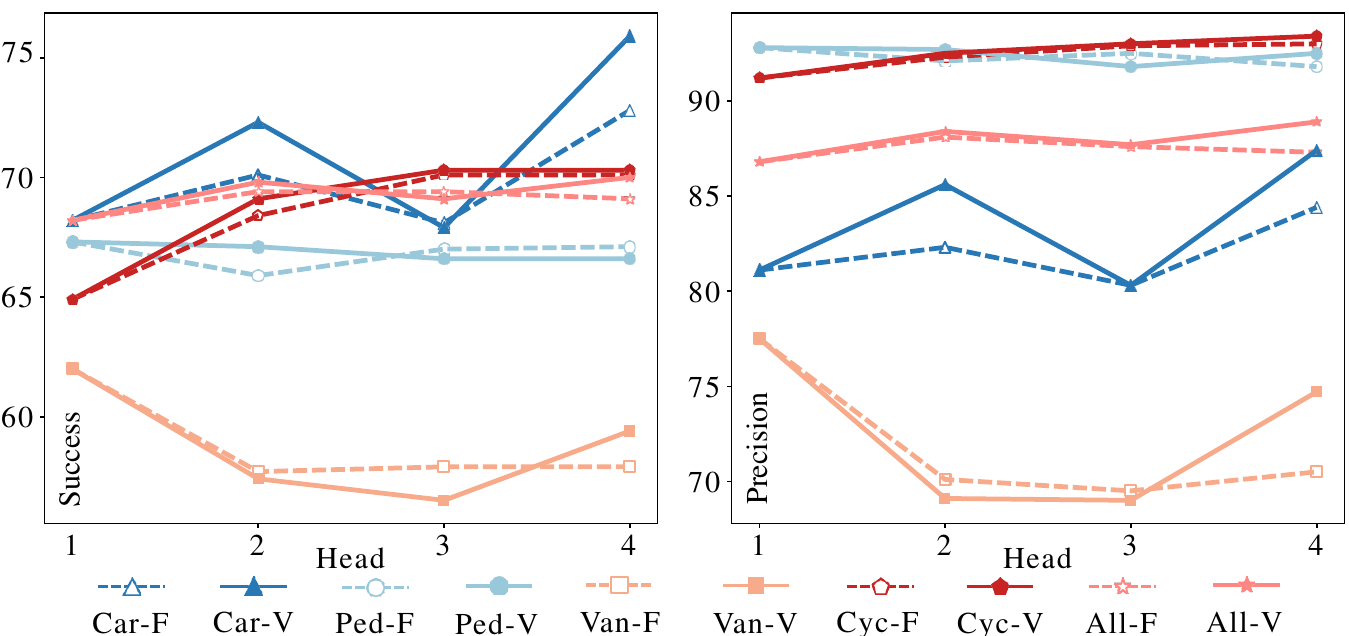}
	\caption{Ablation studies: variable multi-head attention \textit{vs.} fixed multi-head attention.} 
	\label{fig7}
	\vspace{-8pt}	
\end{figure}

\noindent\textbf{SpaceFormer.} For the proposed SpaceFormer using two-stream self-attention, we study the advantages of variable multi-head attention for each category. As shown in Figure \ref{fig7}, compared to fixed multi-head attention, our method is more beneficial to SOT as the number of heads increases.

More results can be found in the supplementary material.

\section{Conclusion}
We discuss a comprehensive framework to serve 3D SOT. The proposed M3SOT consists of multi-frame, multi-field, multi-space, which is a tracking task-oriented pipeline. We analyze the necessity of each module in detail and reveal how to construct tasks to handle the SOT problem. Extensive experiments validate all aspects of the proposed method.

\noindent\textbf{Limitations and Future Work.} We reduce the network load by being task-oriented, however, coordinating such an integrated framework is not easy. We believe that potentially better configurations exist for different scenarios. Moreover, whether to incorporate more components such as motion prediction, multi-target tracking, and how to embed them into the framework are future directions worth exploring.

\section{Acknowledgments}
This work was supported in part by the National Natural Science Foundation of China (62036006, 62276200, 62103314), the Natural Science Basic Research Plan in Shaanxi Province of China (2022JM-327) and the CAAI-Huawei MINDSPORE Academic Open Fund. We acknowledge the support of MindSpore, CANN and Ascend AI Processor used for this research.

\bibliography{references_full}

\newpage
\twocolumn[
\centering{\Large{\textbf{Supplementary Material for \\ M3SOT: Multi-frame, Multi-field, Multi-space 3D Single Object Tracking \\[20pt]}}}
]
	
\section{More Implementation Details}
\subsection{Input Details}
Since the 3D SOT task is oriented to huge scene point clouds with hundreds of thousands of points, the proposed network only needs to consider the subregion in the entire scene where the tracking target may appear for efficiency. Note that unlike previous methods for complex template and search area generation \cite{qi2020p2b}, we dynamically divide template frames and search area frames in time series. Specifically, we scale up the annotated ground-truth bounding box of each point cloud by 2 meters to obtain subregions. We then sample 1024 points respectively within these subregions to generate the input point clouds $\boldsymbol{P^{t-K:t}}$. Furthermore, to deal with inaccurate predictions and enhance robustness to inputs during inference, we perform random translations in the range $[-0.3m, 0.3m]$ in all directions and random rotations around the $z$-axis to augment the input 3D bounding boxes $\boldsymbol{B^{t-K:t-1}}$.

\subsection{Model Details}
We adopt DGCNN \cite{wang2019dynamic} as the backbone network to extract local geometric features, which contains three 1/2 downsampling layers and three EdgeConv layers. In the proposed hybrid transformer, we set up shared MLPs with three linear layers followed by BatchNorm \cite{ioffe2015batch} and ReLU \cite{agarap2018deep} for the various task heads such as mask prediction, center prediction, and bounding box regression. Compared with the similarly designed CXTrack \cite{xu2023cxtrack}, our M3SOT slims down the network, optimizing the network structure and reducing the number of transformations.

\subsection{Training and Inference}
We train our model using the Adam optimizer \cite{kingma2014adam} with an initial learning rate of 0.001. According to the similar experimental phenomenon \cite{lan2022temporal}, the epoch size is set to 50, and the learning rate is reduced to 0.2 every 10 epochs. The batch size is set to 64. We observed that the training for different categories makes the epoch of its convergence different, but it basically fits within 50 epochs. During inference, the model uses the point clouds of the previous two frames to predict the bounding box of the current frame to achieve the tracking objective.

\section{More Experimental Results}
\subsection{All-category Quantitative Results}
Compared to single-category training on the tracking dataset, we try to train all categories and test them together. The advantage of this approach is that the generalization ability of the tracker for all categories can be trained, and the results can also lead to higher and more reliable levels as a whole, as shown in Table \ref{tab8} (a).

We further test the model trained on all categories of KITTI on nuScenes, and obtain results that surpass almost all single categories, see Table \ref{tab8} (b). Note that the corresponding categories between KITTI and nuScenes are Van → Truck and Cyclist → Bicycle. This not only saves the cost of separate experiments for single object tracking (SOT), but also brings inspiration for multiple object tracking (MOT). Since the same scene is annotated with bounding box annotations of multiple categories of single objects, we can choose joint training to not only associate objects with backgrounds, but also associate objects with objects.
\begin{table}[]
	\caption{Comparison of results from full-category training versus single-category training.}
	\vspace{-5pt}
	\label{tab8}
	\centering
	\renewcommand\arraystretch{1.1}
		\begin{tabular}{l|cc}
			\hline
			Category & (a) KITTI  & (b) nuScenes \\
			\hline
			Car $\rightarrow$ Car& {75.9}/87.4 & {34.2}/38.6 \\
			Pedestrian $\rightarrow$ Pedestrian & 66.6/{92.5} & 24.6/37.8 \\
			Van $\rightarrow$ Van (Truck)& 59.4/74.7 & 29.6/25.5 \\
			Cyclist $\rightarrow$ Cyclist (Bicycle) & 56.0/74.0 & 18.8/27.9 \\
			\hline		
			All $\rightarrow$ Car & 72.3/86.0 & 35.0/40.4 \\
			\textit{Improvement} & \textcolor{red}{$\downarrow$3.6}/\textcolor{red}{$\downarrow$1.4} & \textcolor{green}{$\uparrow$0.8}/\textcolor{green}{$\uparrow$1.8} \\
			\hline
			All $\rightarrow$ Pedestrian & 67.7/92.9 & 25.6/39.7 \\
			\textit{Improvement} & \textcolor{green}{$\uparrow$1.1}/\textcolor{red}{$\downarrow$1.4} & \textcolor{green}{$\uparrow$0.8}/\textcolor{green}{$\uparrow$1.8} \\
			\hline
			All $\rightarrow$ Van (Truck) & 68.5/82.3 & 34.3/29.7 \\
			\textit{Improvement} & \textcolor{green}{$\uparrow$9.1}/\textcolor{green}{$\uparrow$7.6} & \textcolor{green}{$\uparrow$4.7}/\textcolor{green}{$\uparrow$4.2} \\
			\hline
			All $\rightarrow$ Cyclist (Bicycle) & 71.0/93.6 & 23.1/39.2 \\	
			\textit{Improvement} & \textcolor{green}{$\uparrow$15.0}/\textcolor{green}{$\uparrow$19.6} & \textcolor{green}{$\uparrow$4.3}/\textcolor{green}{$\uparrow$11.3} \\	
			\hline		
			All $\rightarrow$ All & 70.0/88.9 & 31.7/{38.7} \\
			\hline	
	\end{tabular}
\end{table}
\begin{table*}[]
	\centering
	\caption{Comparison of accuracy-speed tradeoff of different models on KITTI Mean. $*$ represents transformer-based methods.}
	\vspace{-5pt}
	\label{tab9}
	\renewcommand\arraystretch{1.1}
	\resizebox{\linewidth}{!}{
		\begin{tabular}{c|cccccccccc}
			\hline
			Metric & SC3D & P2B & BAT & V2B & M2-Track & PTTR$^*$ & STNet$^*$ & CXTrack$^*$ & MBPTrack$^*$ & M3SOT-F1\&F2$^*$ \\
			\hline
			FPS & 2 & 46 & {56} & 37 & \textbf{57} & 48 & 35 & 29 & 50 & 46\&38\\
			\hline
			Success & 31.2 & 42.4 & 51.2 & 58.4 & 62.9 & 58.4 & 61.3 & 67.5 & \textbf{70.3} & 69.0\&\textbf{70.3}\\
			Precision & 48.5 & 60.0 & 72.8 & 75.2 & 83.4 & 77.8 & 80.1 & 85.3 & 87.9 & 87.0\&\textbf{88.6}\\
			\hline
	\end{tabular}}
	\vspace{-5pt}
\end{table*}
\subsection{Long-range Qualitative Results}
Since Car is the largest category in the tracking dataset, it is also one of the most important targets in the real world. We provide a tracking sequence with a length of 643 frames in Figure \ref{fig10} to evaluate the tracking performance of our M3SOT, which involves various situations such as object loss, density transformation, and orientation transformation. It can be seen from the visualization results that our method can maintain high-accuracy tracking compared with the other state-of-the-art methods, and can make correct transformations to adapt to various abnormal situations.

\section{More Ablations and Analysis}
\subsection{Point Sampling Generation}
We investigate the effect of discrete-state point distributions on target tracking by another efficient random point sampling. As shown as Table \ref{tab10}, it can be seen that the used range sampling all-around outperforms random sampling and can effectively reduce the retrieval time. More, we argue through our pilot study that distance-based farthest point sampling (D-FPS) \cite{qi2017pointnet++} and feature-based farthest point sampling (F-FPS) \cite{yang20203dssd} have significant time-consuming and negative effects on our M3SOT, so we omit reporting on them. We conclude that preserving the local integrity of the input point cloud would provide reliable inspiration for tracking.
\begin{table}[]
	\caption{Ablation studies: point sampling generations.}
	\vspace{-5pt}
	\label{tab10}
	\centering
	\renewcommand\arraystretch{1.1}
	\resizebox{\linewidth}{!}{
		\begin{tabular}{c|cccc|c}
			\hline
			Method &  Car& Pedestrian& Van&  Cyclist& All\\
			\hline
			Random &  70.3/82.3 & 62.3/88.1 & 57.5/70.3&  69.8/92.6 & 66.0/84.3\\
			\textbf{Range} &  \textbf{75.9}/\textbf{87.4} & \textbf{66.6}/\textbf{92.5} & \textbf{59.4}/\textbf{74.7} & \textbf{70.3}/\textbf{93.4} & \textbf{70.0}/\textbf{88.9}\\
			\hline	
	\end{tabular}}
	\vspace{-5pt}
\end{table}

\subsection{Visual Analysis via Multi-Frame Input}
With multi-frame input as our core design, we further analyze the effect of template set size on the search area. As shown in Figure \ref{fig9}, we analyze the response degree of the four categories search area point clouds ($\boldsymbol{P^t}$) to different numbers of input point clouds, respectively. The attention maps show that when the template set size is 2, the search area can be clearly distinguished from the targeting information. A phenomenon is that for large-scale targets (\textit{e.g.}, Car and Van), our M3SOT has difficulty distinguishing perfect fine-grained information, but can still focus on the central region of objects for robust regression to bounding boxes.
\begin{figure}[t]
	\centering
	\includegraphics[width=\linewidth]{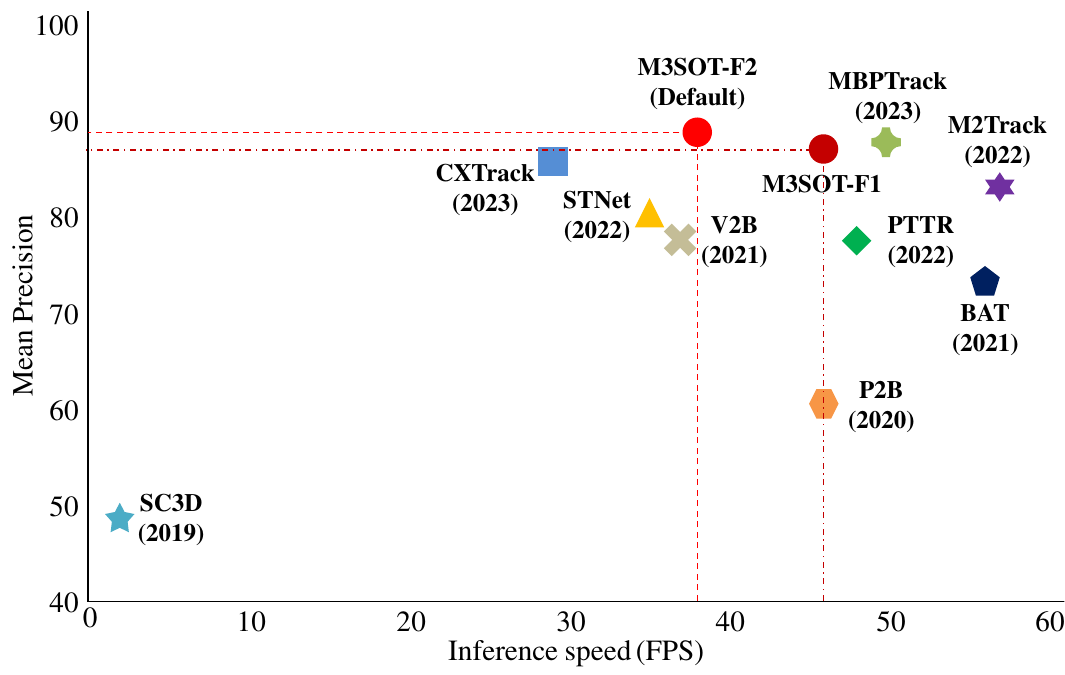}
	\caption{Accuracy-speed tradeoff on KITTI Mean. Our M3SOT performs best. } 
	\label{fig8}
\end{figure}
\begin{figure}[t]
	\centering
	\includegraphics[width=\linewidth]{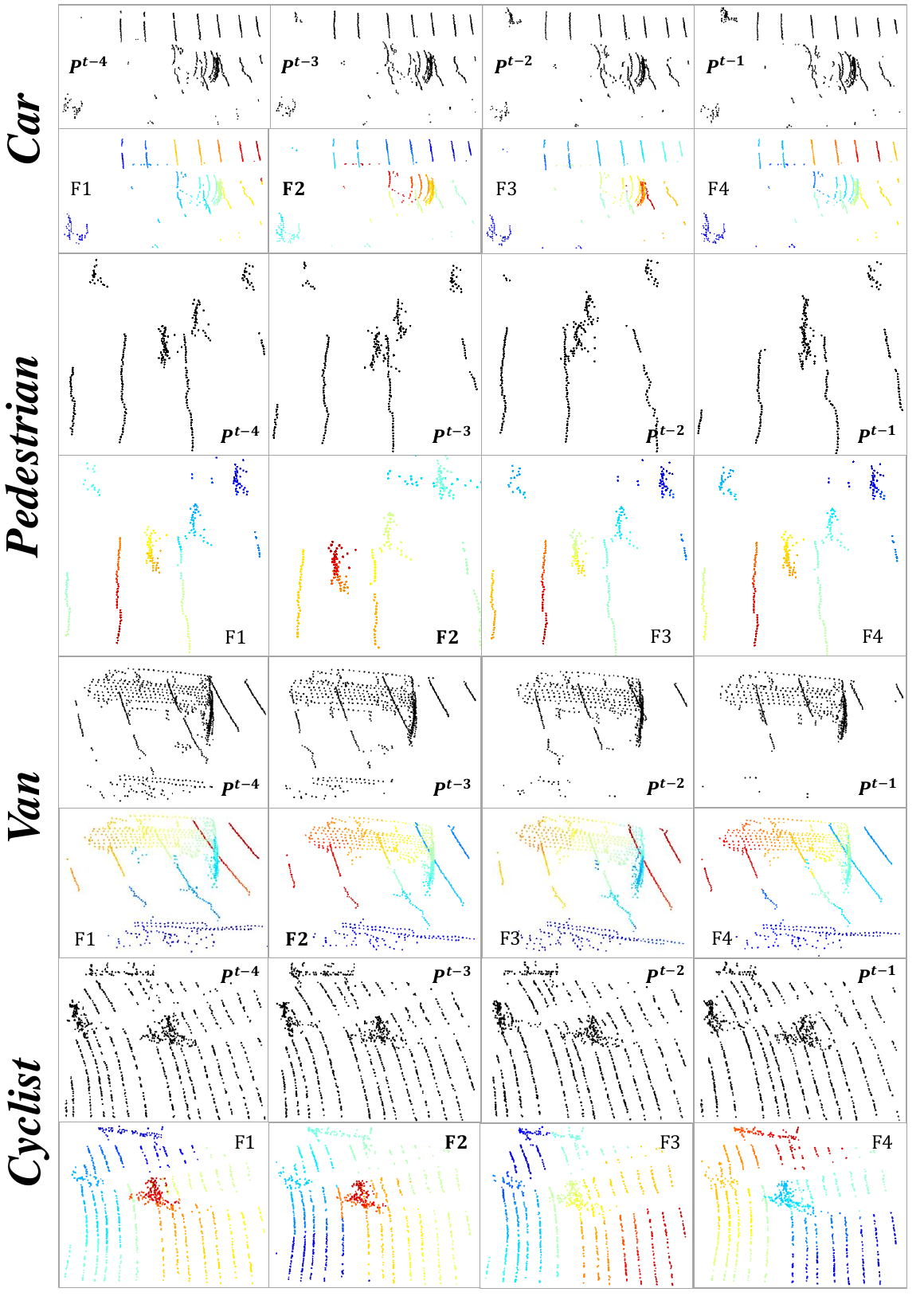}	
	\caption{The effect of template set size on the search area from different categories.} 
	\label{fig9}
\end{figure}

\subsection{Inference Speed}
Since speed is a critical factor in 3D SOT, we validate the inference speed of the trained model using a single GeForce RTX 3090 GPU. For the average frame, the M3SOT model takes 4.2 ms for pre/post-process, 8.8 ms for backbone, 10.8 ms for forward propagation, and 2.4 ms for localization head. Compared with the transformer-based methods, M3SOT requires extra inference time to handle multiple input frames. Note that we also report M3SOT-F1 without extra frames. As a result, we report the Mean Success/Precision and FPS, as shown in Table \ref{tab9} and Figure \ref{fig8}. Overall, our method achieves superior results at an acceptable cost.
\begin{figure*}[]
	\centering
	\includegraphics[width=\linewidth]{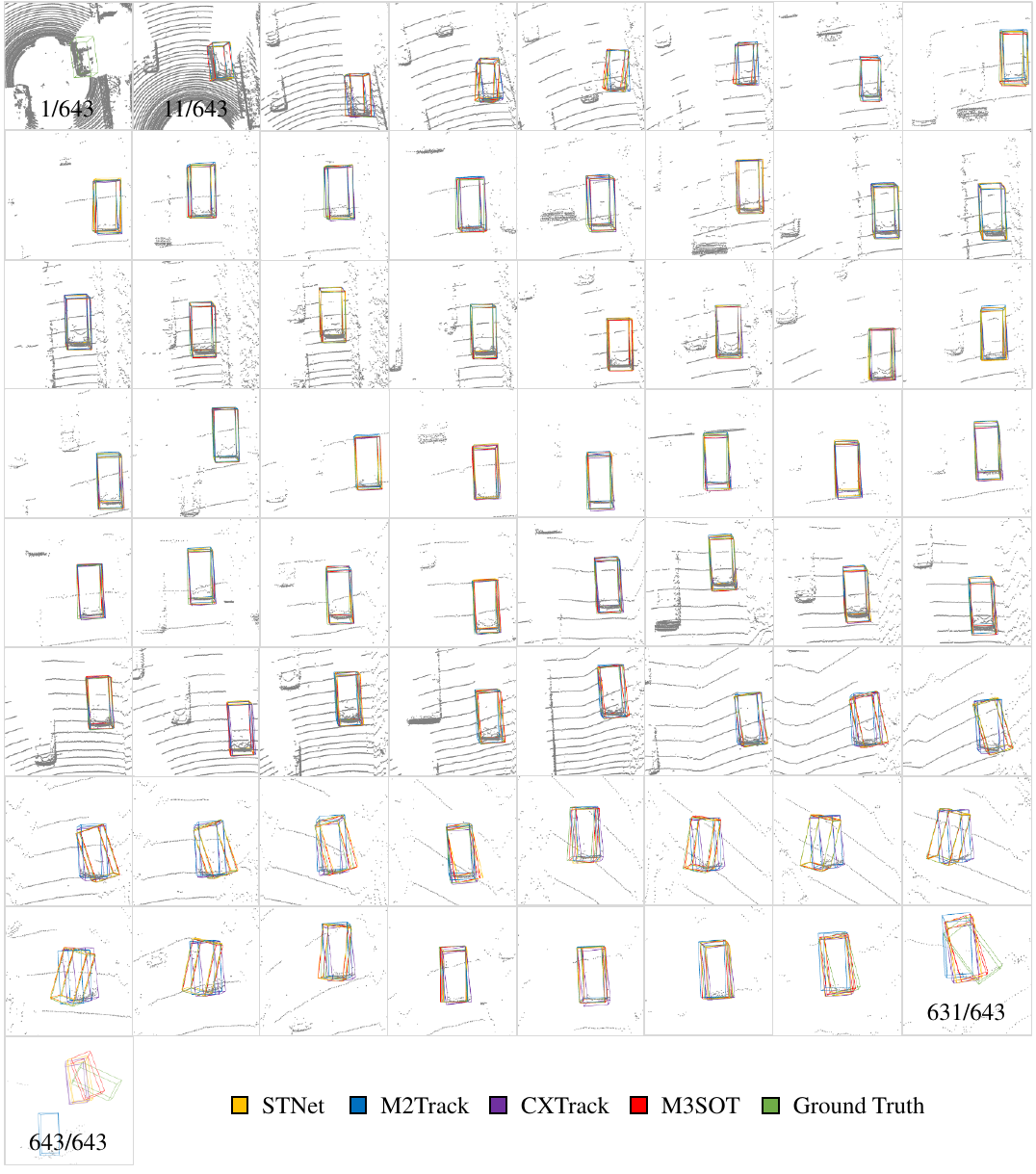}
	\vspace{-5pt}
	\caption{Visualization of long-range tracking results from different methods on KITTI. A scene is switched every 10 frames.} 
	\label{fig10}
	\vspace{-5pt}
\end{figure*}
	

\end{document}